\documentclass[a4paper]{article}

\usepackage{color}
\usepackage{multirow}
\usepackage{bm}
\usepackage{algorithm}  
\usepackage{algorithmicx}  
\usepackage{algpseudocode}  

\usepackage{INTERSPEECH2018}

\title{Training Augmentation with Adversarial Examples\\ for Robust Speech Recognition}

\name{Sining Sun$^1$ 
,  Ching-Feng Yeh$^2$,  Mari Ostendorf$^3$, Mei-Yuh Hwang$^2$, Lei Xie$^{1*}$ \thanks{The research work is supported by the National Key Research and Development Program of China (Grant No.2017YFB1002102) and the National Natural Science Foundation of China (Grant No.61571363).}\thanks{$^*$ Lei Xie is the corresponding author. }}

\address{
  $^1$School of Computer Science, Northwestern Polytechnical University, Xi'an, China\\
  $^2$Mobvoi AI Lab, Seattle, USA\\
  $^3$Department of Electrical Engineering, University of Washington, Seattle , USA}
\email{\{snsun,lxie\}@nwpu-aslp.org, \{cfyeh,mhwang\}@mobvoi.com, ostendor@uw.edu}

\begin{document}

\maketitle
%

\begin{abstract}
This paper explores the use of adversarial examples in training  speech recognition systems to increase robustness of deep neural network acoustic models.
During training, the fast gradient sign method is used to generate adversarial examples 
augmenting the original training data. 
Different from conventional data augmentation based on data transformations, the examples are dynamically generated based on current acoustic model parameters.
We assess the impact of adversarial data augmentation in experiments on the Aurora-4 and CHiME-4 single-channel tasks, showing improved robustness against noise and channel variation.
Further improvement is obtained when combining adversarial examples with teacher/student training, leading to a 23\% relative word error rate reduction on Aurora-4.
\end{abstract}

\section{Introduction}
\label{sec1_intro}
In recent few years, there has been significant progress in automatic speech recognition (ASR) due to the successful application of deep neural networks (DNNs)~\cite{dahl2012context,hinton2012deep}, such as  convolutional neural networks (CNNs)~\cite{abdel2012applying, sainath2013deep}, recurrent neural networks (RNNs)~\cite{graves2013speech} and sequence-to-sequence learning~\cite{graves2013speech}. Acoustic modeling based on deep learning has shown robustness against noisy signals due to the deep structure and an ability to model non-linear transformations~\cite{qian2016very}.
However, current ASR systems are still sensitive to environmental noise, room reverberation~\cite{kinoshita2013reverb} and channel distortion. 
A variety of methods have been proposed to deal with noise, including front-end processing such as data augmentation~\cite{ko2015audio}, single or multi-channel speech enhancement~\cite{xiao2016study}, robust feature transformation, very deep CNNs~\cite{rennie2014deep,qian2016very}, and learning strategies such as teacher-student (T/S)~\cite{li2017large} and adversarial training~\cite{SUN201779}. Here, we focus on data augmentation, assessed with and without T/S training.

The performance of acoustic models degrades when there exists a mismatch between training data and unseen test data. Data augmentation is an effective way to improve robustness, aiming to increase the diversity of the training data by augmenting it with perturbed versions using methods such as adding noises or reverberation to clean speech. Acoustic modeling with  data augmentation is also known as multi-condition or multi-style training~\cite{lippmann1987multi}, which has been widely adopted in many ASR systems.
Data augmentation strategies for reverberant signals was investigated in~\cite{ko2015audio}, which proposed using real and simulated room impulse responses to modify clean speech with different signal-to-noise ratio (SNR) levels. Generally, the robustness of the system improves with the diversity in the training data. 
A variational autoencoder (VAE) approach to data augmentation  was proposed in~\cite{hsu2017unsupervised} for unsupervised domain adaptation. They train a VAE on both clean and noisy speech without supervised information to learn a latent representation. The noisy data to be augmented are then selected by comparing the similarities with the original data, with latent representations modified by removing attributes such as speaker identity, channel and background noise.

Teacher-student (T/S) training has been widely adopted as an effective approach to increase the robustness of acoustic modeling in supervised~\cite{watanabe2017student} or unsupervised~\cite{li2017large} scenarios.  T/S training relies on parallel data to train a teacher model and a student model. For example, a close-talk data set can be used to train the teacher model, while the same speech collected by a far-field microphone can be used to train the student model. Adversarial training, which aims to learn a domain-invariant representation, is recently proposed for robust acoustic modeling~\cite{sundomain, shinohara2016adversarial}. Adversarial training can be used to reduce the mismatches between training and test data, and it is applicable in both supervised and unsupervised scenarios.

In this paper, we propose data augmentation with adversarial examples for acoustic modeling, in order to improve robustness in adverse environments. The concept of adversarial examples was first proposed in~\cite{szegedy2013intriguing} for computer vision tasks. They discovered that neural networks can easily misclassify examples in which the image pixels are only slightly skewed from the original ones. That is, the models can be very sensitive to even minor input disturbance.
Adversarial examples have provoked research interest in computer vision and natural language processing~\cite{43405,jia2017adversarial}. Recently, adversarial examples were introduced to simulate attacks to state-of-the-art end-to-end ASR systems~\cite{carlini2018audio}. In the work, a white-box targeted attack scenario was shown: given a natural waveform $x$ and a nearly inaudible adversarial noise $\delta$ which is generated from $x$ and some targeted phrase $y$, $x+\delta$ would be recognized as $y$ regardless of the original content in $x$. This is consistent with the observation in~\cite{szegedy2013intriguing} where neural networks can be vulnerable when there are minor but elaborate disturbances.

Previous work has focused on improving model robustness against adversarial test examples~\cite{43405,kurakin2016adversarial}. In our work, we adopt the idea and augment training data with adversarial examples to obtain more robust acoustic models to natural data instead of adversarial examples only. 
In contrast to adversarial training, where the model is trained to be invariant to specific phenomena represented in the training set, the adversarial examples used here are generated automatically based on inputs and model parameters associated with each mini-batch.
In the training stage, we generate adversarial examples dynamically using the fast gradient sign method (FGSM)~\cite{43405}, since it has been shown to be both effective and efficient compared with other approaches for generating adversarial examples. For each mini-batch, after the adversarial examples are obtained, the parameters in the model are updated with both the original and the adversarial examples. Furthermore, we combine the proposed data augmentation scheme with teacher-student (T/S) training when parallel data is available, and find that the improvements from both approaches are additive. 

The rest of the paper is organized as follows: Sec.~\ref{sec2_advex} introduces adversarial examples and how to generate them using FGSM. Sec.~\ref{sec3} gives details of using adversarial examples for acoustic modeling. Sec.~\ref{sec4_exp} describes the experimental setup and results on the CHiME-4 single track tasks and on the Aurora-4 dataset. Concluding remarks are presented in Sec.~\ref{sec5}.

\section{Adversarial examples}
\label{sec2_advex}
\subsection{Definition of adversarial examples}
The goal of adversarial examples is to 
disturb well-trained machine learning models. Relevant work shows that state-of-the-art models can be vulnerable to adversarial examples; i.e., the predictions of the models are easily misled by non-random perturbation on input signals, even though the perturbation is hardly perceptible by humans~\cite{szegedy2013intriguing}. In such cases, these perturbed input signals are carefully designed and named ``adversarial examples.'' The success of using adversarial examples to disturb models also indicates that the output distribution of neural networks may not be smooth with respect to the instances of input data distribution. As a result, a small skew in the input signals may cause abrupt changes on the output values of the models~\cite{miyato2015distributional}.  

In general, a machine learning model, such as a neural network, is a parameterized function, $f(\bm{x};\bm{\theta})$, where $\bm{x}$ is the input and $\bm{\theta}$ represents the model's parameters. A trained model $f(\bm{x};\bm{\theta})$ is used to predict the label ${y_i}$ given the input $\bm{x}_i$. An  adversarial example $\bm{x}_{i}^{adv}$ can be constructed as:
\begin{equation}\label{eq1}
\bm{x}_{i}^{adv}=\bm{x}_i +\bm{\delta}_i
\end{equation}
so that
\begin{equation}
y_i\neq f(\bm{x}_{i}^{adv};\bm{\theta})
\end{equation}
where 
\begin{equation}\label{eq3}
\|\bm{\delta}_i\| \ll \|\bm{x}_i\| ,
\end{equation}
and $\bm{\delta}$ is called the adversarial perturbation.
For a trained and robust model, small random perturbations should not have a
significant impact on the output of the model. Therefore, generating adversarial perturbations as negative training examples can potentially improve model robustness.

\subsection{Generating adversarial examples}
In ~\cite{43405}, the fast gradient sign method (FGSM) was proposed to generate adversarial examples using current model parameters and existing training data to generate adversarial perturbations $\bm{\delta}_i$ in equation~\ref{eq1}.

Given model parameters $\bm{\theta}$, inputs $\bm{x}$ and the targets $y$ associated with $\bm{x}$, the model is trained to minimize the loss function $J(\bm{\theta}, \bm{x}, y)$. In this work, we use average cross-entropy for $J(\bm{\theta}, \bm{x}, y)$, which is very common for classification tasks. Conventionally, to train a neural network, gradients are computed with the predictions of the model and the designated label, and the gradients are propagated through the layers using the back-propagation algorithm 
until the input layer of the network is reached. However, it is possible to further compute the gradient with respect to the input to the network (to find adversarial examples) rather than just the network weights~\cite{43405}. 


The idea of FGSM is to generate adversarial examples that maximize the loss function $J(\bm{\theta}, \bm{x}, y)$,
\begin{eqnarray} \label{eq5}
	\bm{x}^{adv} & =  & \mathop{\arg\max}_{\bm{x}} \      {J(\bm{\theta}, \bm{x}, y)} \\
    & = & \bm{x} + \bm{\delta}_{FGSM} ,
\end{eqnarray}
where
\begin{equation} \label{eq4}
\bm{\delta}_{FGSM}=\epsilon \,\mbox{sign}({\nabla}_{\bm{x}}{J(\bm{\theta}, \bm{x}, y)})
\end{equation}
and $\epsilon$ is a small constant to be tuned. Note that FGSM uses the sign of the gradient instead of the value, making it easier to satisfy the constraint of equation~\ref{eq3}. Our experimental results show that small $\epsilon$ is stable to generate adversarial examples to perturb the neural networks.

\section{Training with adversarial examples}
\label{sec3}

Different from other data augmentation approaches such as adding artificial noises to simulate perturbations, the adversarial examples are generated by the model but shifted to a bigger loss value ${\nabla}_{\bm{x}^{adv}}{J(\bm{\theta}, \bm{x}^{adv}, y)}$ than the original data $\bm{x}$. After $\bm{x}^{adv}$ is generated, we further use it to update the model parameters, in order to enhance the robustness of the ASR system against noisy environments. In this work, FGSM is used to generate adversarial examples dynamically within each mini-batch, and the model parameters are updated immediately following the original mini-batch, as elaborated in Algorithm 1.

\begin{algorithm} 
\caption{Training neural network with automatically generated adversarial examples}
\begin{algorithmic}[1] 
\Require

	{$D=\{\bm{x}_i, y_i\}_{k=1}^{K}$, training set 
    
    $\bm{x}_i$, input features
    
    $y_i$, output labels
    
    $\mu$, learning rate 
    
    $\epsilon$, adversarial weight }
    
\Ensure 
	{$\bm{\theta}$, model parameters}
    
\   
\State Initialize model parameters $\bm{\theta}$
\While {model does not converge}
	\State Read a mini-batch $B=\{\bm{x}_m, y_m\}_{m=1}^M$ from $D$
    \State Train model using $B$, 
    
    \
    $\bm{\theta}\gets\bm{\theta} -  \frac{\mu}{M}\sum_{m=1}^{M}{{\nabla}_{\bm{\theta}}J(\bm{x}_m, y_m,\bm{\theta})} $
    \State Calculate $\{\bm{\delta}_m^{FGSM}\}_{m=1}^M$ using equation \ref{eq4} for $B$

\
$\bm{\delta}_{m}^{FGSM}=\epsilon \,\mbox{sign}({\nabla}_{\bm{x}_m}{J(\bm{x}_m, y_m,\bm{\theta})})$

    \State Generate adversarial examples using equation~\ref{eq1}
    
    \ 
    $\bm{x}_m^{adv}=\bm{x}_m + \bm{\delta}_{m}^{FGSM}$
    
    \State Make a mini-batch $B_{adv}$ with $\{\bm{x}_{m}^{adv}\}_{m=1}^{M}$
    
    \ 
    $B_{adv}=\{\bm{x}_{m}^{adv}, y_m\}_{m=1}^M$
    \State Train model using $B_{adv}$, 
    
    \
    $\bm{\theta}\gets\bm{\theta} -  \frac{\mu}{M}\sum_{m=1}^{M}{{\nabla}_{\bm{\theta}}J(\bm{x}_m^{adv}, y_m,\bm{\theta})} $
     
\EndWhile
\end{algorithmic}
\end{algorithm}

In the case of acoustic modeling, the input $\bm{x}$ refers to acoustic features such as Mel-frequency cepstral coefficients (MFCCs), the label $y$ refers to frame-level alignments such as indices of senones from forced alignments. With data augmentation using adversarial examples, we apply several steps with each mini-batch to update the model parameters: (1) Train the model with the original inputs and obtain the adversarial perturbations $\bm{\delta}_m$ per sample as in equation~\ref{eq4}. As $\epsilon$ is a constant, the perturbation for each feature dimension of each sample is either $+\epsilon$ or $-\epsilon$. (2) Generate adversarial examples with the obtained adversarial perturbations. (3) Make a mini-batch with adversarial examples and original labels. (4) Train the model with the mini-batch from the adversarial examples. 

By feeding the original labels in the mini-batch with adversarial examples, we are minimizing the distance between the ground truth and output of the network as regular training. But what is worth noticing is that the adversarial examples are generated by FGSM to maximize the loss function as described in equation~\ref{eq5}, which reflects the ``blind spots" of the current model in input space to some extent.

\section{Experiments}
\label{sec4_exp}

\subsection{Speech corpora and system description }
\label{sec4.1}
\subsubsection{Aurora-4 corpus}
\label{secAurora}
The Aurora-4 corpus is designed to evaluate the robustness of ASR systems on a medium vocabulary continuous speech recognition task based on Wall Street Journal (WSJ0, LDC93S6A). There are 7138 clean utterances from 83 speakers in the training set, recorded using the primary microphone, denoted as WSJ0 corpus here. The multi-condition training set (denoted as WSJ0m) also consists of 7138 utterances, but with a combination of clean and noisy speech perturbed by one of six different noises at 10-20 dB SNR. 
Channel distortion is introduced by recording the data with two different microphones. The test data set consists of four subsets: clean, noisy, clean with channel distortion, and noisy with channel distortion. The noise is simulated. The four test sets
are referred to as A, B, C, D respectively in the literature. There are 330 utterances in test set A and C, and 1980 ($330\times 6$) utterances in B and D respectively. 
For model tuning, we use a 330-utterance subset (dev\_0330)~\cite{pearce2002aurora} of the official 1206 utterance dev set as our development set, for reasons that will be explained later. More details about Aurora-4 corpus can be found in~\cite{hirsch2000aurora}.

\subsubsection{CHiME-4 corpus}
The CHiME-4 task is a speech recognition challenge for single-microphone or multi-microphone tablet device recordings in everyday scenarios under noisy environments. For the CHiME-4 data set, there are four noisy recording environments: street (STR), pedestrian area (PED), cafe (CAF) and bus (BUS). For training, 1600 utterances were recorded in the four noisy environments from four speakers, and additional 7138 noisy utterances simulated from WSJ0 by
additive noises from the four noisy environments. 
The development set consists of 410 utterances in each of the four environments with both real (dt05\_real) and simulated environments (dt05\_simu), for a total of 3280 utterances.  There are 2640 utterances in the evaluation set, with 330 utterances in each of the same eight conditions.

\subsubsection{System description}
\label{set4.2}
We adopt convolutional neural networks (CNNs) for acoustic modeling for all the experiments we report in this work. The configurations for our CNNs are consistent with the previous work in~\cite{rennie2014deep}, which has two convolutional layers with 256 feature maps in each layer. $9\times 9$ filters with $1\times 3$ pooling is used in the first layer and $3\times 4$ filters in the second layer without pooling. There are four fully-connected layers with 1024 hidden units after the convolutional layers. Rectified linear unit (ReLU) activation function is used for all layers. For the Aurora-4 setup, 40-dimensional mel-filter bank (fbank) features with 11-frame context window are used as inputs. For the CHiME-4 setup, 40-dimensional fMLLR features with 11-frame context window are used. 
Standard recipes in Kaldi~\cite{povey2011kaldi} are adopted for feature extraction, HMM-GMM training and alignment generation. As for acoustic modeling, TensorFlow~\cite{abadi2016tensorflow}
is used for training CNNs in this work with cross-entropy as the objective function and Adam~\cite{kingma2014adam}
as the optimizer. The sizes for output layers are 2025 and 1942 for the Aurora-4 and CHiME-4 tasks, respectively. 

As Aurora dev\_0330 contains verbalized
punctuations, we use the 20k closed-vocab bigram LM with verbalized punctuations during Aurora development, to fine-tune our hyper-parameter $\epsilon$ and LM weight. This dev subset
is chosen so that all words are covered by the 20k vocabulary, 
to avoid confounding effects of noise with out-of-vocabulary issues.
For the evaluation sets, we use
the official WSJ 5k closed vocabulary with a 3-gram model with non-verbalized punctuations (5c-nvp 3gram), since the 5k vocabulary covers all words
in the test sets.

The values of $\epsilon$ and the language weight for both tasks  are tuned on the respective development sets, and the best hyper-parameters are then applied to the evaluation sets. 

\subsection{Experimental results}
\begin{figure}[!htbp]
\includegraphics[width=0.41\textwidth]{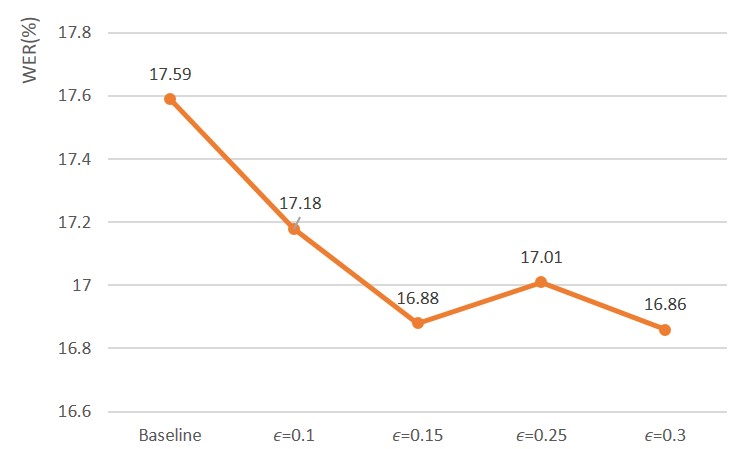}
\caption{WER on Aurora-4 dev\_0330 with adversarial data augmentation using various perturbation weights.}
\label{fig1}
\end{figure}

\begin{table}[!htbp]
\centering
\caption{WER comparison on the Aurora-4 evaluation set with adversarial examples (AdvEx) ($\epsilon=0.3$)}
\label{tab1}
\begin{tabular}{|c|c|c|c|c|c|}
\hline
       & A    & B    & C    & D     & AVG.          \\ \hline
Baseline     & 3.21 & 6.08 & 6.41 & 18.11 & 11.05         \\ \hline
AdvEx  & 3.51 & 5.84 & 5.79 & 14.75 & 9.49          \\ \hline
WER reduction (\%) & -9.4 & 3.9 & 9.7 & 18.6 & 14.1 \\ \hline
\end{tabular}
\end{table}

\begin{table*}[tbp]
\centering
\caption{WER comparison on CHiME-4 single-channel track evaluation sets with adversarial examples (AdvEx) ($\epsilon=0.1$).}
\label{table_chime4}
\begin{tabular}{|c|c|c|c|c|c|c|c|c|c|c|}
\hline
\multirow{2}{*}{system} & \multicolumn{5}{c|}{et05\_simu}                & \multicolumn{5}{c|}{et05\_real}                \\ \cline{2-11} 
                        & BUS   & CAF   & PED   & STR   & \textbf{AVE.}  & BUS   & CAF   & PED   & STR   & \textbf{AVE.}  \\ \hline
Baseline  & 20.25 & 30.69 & 26.62 & 28.74 & 26.57          & 43.95 & 33.64 & 25.95 & 18.68 & 30.55          \\ \hline
AdvEx                  & 19.65 & 29.29 & 24.75 & 26.95 & \textbf{25.16} & 41.00 & 31.34 & 24.74 & 18.23 & \textbf{28.82} \\ \hline
\end{tabular}
\end{table*}

\subsubsection{Aurora-4 results}
Figure~\ref{fig1} shows the word error rate (WER) results on dev\_0330 for different $\epsilon$. Based on the results, $\epsilon=0.3$ is chosen as the best perturbation weight to train the Aurora-4 model. Table~\ref{tab1} shows the results on the Aurora-4 evaluation set. The model trained on WSJ0m serves as the baseline. With $\epsilon=0.3$, the augmented data training achieves 9.49\% WER averaged across the four test sets, a 14.1\% relative improvement over the baseline. For the test set with the highest WER on the baseline system, D, in which both noise and channel distortion are present, the proposed method reduced the WER by 18.6\% relative. We also experimented with dropout training, and it gave only very small gains over the baseline.

\subsubsection{CHiME-4 results}
\begin{figure}
\includegraphics[width=0.43\textwidth]{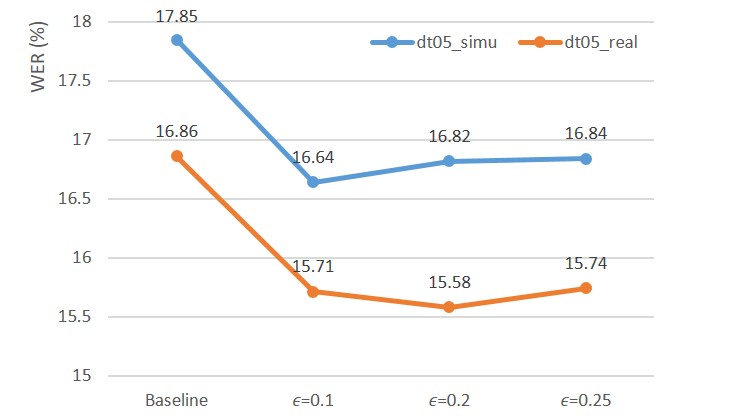}
\caption{WER with adversarial data augmentation on CHiME-4 
dt05\_simu and dt05\_real data sets,  with various  perturbation weights.}
\label{fig2}
\end{figure}

In Figure~\ref{fig2}, different values of $\epsilon$ are selected to demonstrate the impact of $\epsilon$ on dt05\_simu and dt05\_real sets on CHiME-4. We see that within a reasonable range ($\epsilon< 0.25$) the proposed approach brings consistent gain. 
In Table~\ref{table_chime4}, results on CHiME-4 single track are listed, including the real (et05\_real) and simulated (et05\_simu) evaluation sets. Relative WER reductions obtained on et05\_real and et05\_simu sets were 5.7\% and 5.3\%, respectively. 
The proposed approach was able to bring consistent improvements for all types of noises, whether in simulated or real environments.

\subsubsection{Combining T/S training with data augmentation}

Teacher-student (T/S) training
has proven to be effective to improve the robustness of the model in scenarios where parallel data is available. As a result, in this work we also tried to combine T/S training with the proposed data augmentation method. As described in Section~\ref{secAurora}, parallel training data is available for  Aurora-4. Accordingly, a teacher model is trained using clean data, while the noisy data  is used to train the student model. While training the student model, the following loss function is used to optimize the parameters:
\begin{equation}
\begin{aligned}
J_{T/S} &= \alpha\, {\mbox CE}(y,f(\bm{x}_n,\bm{\theta}_S))
\\&+(1-\alpha)\,{\mbox CE}(y_T,f(\bm{x}_n,\bm{\theta}_S))
\end{aligned}
\end{equation}
where $0<\alpha<1$ is the discount weight, CE refers to the cross-entropy loss, $y$ is the posterior probability estimated by the student model, $\bm{x}_n$ is a noisy (or adversarial) example, $\bm{\theta}_S$ is the student model parameter, and $y_T$ is the posterior probability estimated from the teacher model using clean data $\bm{x}_c$, 
\begin{equation}
y_T = f(\bm{x}_c, \bm{\theta}_T)
\end{equation}
where $\bm{\theta}_T$ is the teacher model. The teacher model has the same configuration as the student model. This learning strategy 
uses a similar loss function as Kullback–Leibler divergence regularization~\cite{Yu2013}.

Table~\ref{tab3} shows the WER of T/S learning with $\alpha=0.5$. T/S learning alone gives us $12.2\%$ relative WER reduction. After combining with adversarial data augmentation, we get the best performance with 8.50\% WER. 

\subsubsection{Random perturbations}
In order to assess the utility of data augmentation using adversarial examples, we compare the proposed approach with data augmentation using random perturbation instead of FSGM. For random perturbation, we replace $\mbox{sign}({\nabla}_{\bm{x}}J(\bm{\theta}, \bm{x}, y))$ in equation~\ref{eq4} with a random $\pm 1$ value. The last row of Table~\ref{tab3} shows that augmenting data using random perturbation gives little gain (T/S+Random) compared to using T/S learning alone. However, there is a significant performance gap between  the two data augmentation methods, even though the sizes of the augmented training data are the same. This verifies the effectiveness of adversarial examples for robust acoustic modeling. 

\begin{table}[]
\centering
\caption{WER when combining T/S with adversarial data augmentation on Aurora-4. }
\label{tab3}
\scalebox{0.96}[0.96]{
\begin{tabular}{|c|c|c|c|c|c|}
\hline
                                        & A    & B    & C    & D     & AVG.          \\ \hline                                     
Baseline     & 3.21 & 6.08 & 6.41 & 18.11 & 11.05         \\ \hline
T/S($\alpha$=0.5)                       & 2.86 & 5.49 & 5.25 & 15.80  & 9.70           \\ \hline
T/S+AdvEx($\epsilon$=0.3)  & 3.08 & 5.42 & 4.89 & 13.09 & \textbf{8.50}           \\ \hline
T/S+Random($\epsilon$=0.3) & 3.62 & 5.69 & 5.60  & 14.89 & 9.48          \\ \hline
\end{tabular}
}
\end{table}

\section{Conclusions}
\label{sec5}

In this work, we propose data augmentation using adversarial examples for robust acoustic modeling. During training, FGSM is used to efficiently generate adversarial examples, in order to increase the diversity of the training data. Experimental results on Aurora-4 and CHiME-4 tasks show that the proposed approach can improve the robustness of acoustic modeling with deep neural networks against noise and channel variation. On the Aurora-4 evaluation set, 14.1\% relative WER reduction was obtained, with the greatest benefit (18.6\%) when both noise and channel distortion are present. On the CHiME-4 single track task, roughly 5\% WER reductions were obtained on both real and simulated data. Similar to the use of simulated data, adversarial examples effectively increase the size of the training set without actually requiring new data. These results show that the methods are useful in combination. Adding teacher-student learning further improved performance on the Aurora-4 task, leading 23\% relative WER reduction overall. 
Training with adversarial examples is similar in spirit to discriminative training; it would be interesting to compare and combine these approaches with different size training sets. This finding suggests that the use of adversarial examples for data augmentation is likely to be complementary to other methods for improving robustness, offering opportunities for future work. 



\clearpage
\bibliographystyle{IEEEtran}

\bibliography{main}


\end{document}